\documentclass[letterpaper, 10 pt, conference]{ieeeconf}  
\IEEEoverridecommandlockouts                         
\usepackage[utf8]{inputenc}
\usepackage{hyperref}
\usepackage{graphicx}
\usepackage[margin=1in]{geometry}
\input{mysymbol.sty}
\usepackage{amsmath}
\usepackage{amssymb}
\usepackage{xcolor}
\usepackage{subfigure}
\usepackage{pgfplots}

\newlength\figureheight
\newlength\figurewidth

\newcommand\copyrighttext{%
  \footnotesize \textcopyright 2022 IEEE. Personal use of this material is permitted.
  Permission from IEEE must be obtained for all other uses, in any current or future 
  media, including reprinting/republishing this material for advertising or promotional 
  purposes, creating new collective works, for resale or redistribution to servers or 
  lists, or reuse of any copyrighted component of this work in other works.
  
  This paper has been accepted for publication at the $39$th IEEE International Conference on Robotics and Automation (ICRA 2022).}
\newcommand\copyrightnotice{%
\begin{tikzpicture}[remember picture,overlay]
\node[anchor=south,yshift=10pt] at (current page.south) {\fbox{\parbox{\dimexpr\textwidth-\fboxsep-\fboxrule\relax}{\copyrighttext}}};
\end{tikzpicture}%
}

\title{Optimizing Camera Placements for Overlapped Coverage with 3D Camera Projections}
\author{Akshay Malhotra$^{1}$, Dhananjay Singh$^2$, Tushar Dadlani$^3$, Luis Yoichi Morales$^{2}$
\thanks{$^{1}$InterDigital Comm.,
        {\tt\small akshaymalhotra190@gmail.com}}%
\thanks{$^{2}$Standard Cognition,
        {\tt\small \{dj,luis\}@standard.ai}}%
        \thanks{$^{3}$Walmart,
        {\tt\small tushardadlani@gmail.com}}
\thanks{This work was completed at Standard Cognition.}
}

\pgfplotsset{compat=1.16}

\begin{document}

\maketitle

\thispagestyle{empty}
\pagestyle{empty}
\copyrightnotice
\begin{abstract}
This paper proposes a method to compute camera $6$ $DoF$ poses to achieve a user defined coverage. 
The camera placement problem is modeled as a combinatorial optimization where given the maximum number of cameras, a camera set is selected from a larger pool of possible camera poses. We propose to minimize the squared error between the desired and the achieved coverage, and formulate the non-linear cost function as a mixed integer linear programming problem. 
A camera lens model is utilized to project the camera's view on a 3D voxel map to compute a coverage score which makes the optimization problem in real environments tractable.
Experimental results in two real retail store environments demonstrate the better performance of the proposed formulation in terms of coverage and overlap for triangulation compared to existing methods.

\end{abstract}

\section{introduction}
\label{intro}
Development of internet of things device technology, increased network speeds and reduced sensor costs have revolutionized the field of network connected systems \cite{wang2011coverage} with application to surveillance \cite{Shen2007AMS}, people flow detection and behavior analysis \cite{brvsvcic2013person}. 
Such systems use a multitude of sensors such as cameras, LiDAR's, or depth sensors to cover the environment of interest. 
One of the main challenges for these systems is the computation of the sensor poses to ensure environmental coverage. 

This problem has been studied in the field of computational geometry and is known as the art gallery problem \cite{urrutia2000art}.
The art gallery problem computes the number of guards or cameras necessary to observe or cover the environment which is usually solved as a geometrical problem in two dimensions. Camera coverage computation in real environments is a  three dimensional problem that still has not been widely discussed.


Development of self and automated checkout systems \cite{amazongo, zippin, liu2020grab, ruiz2019aim3s} and recent interest for the automation of warehouse shelf inventory 
have increased the interest for sensor placement algorithms in real environments.
For these systems it is necessary to perform sensor placement considering different types of constraints. For instance, for a surveillance system, it would be of interest to have good views of regions of interest such as products on shelves while allowing human body and hand(s) detection.  
Top-down cameras could cope with the problem of occlusions for human detection and tracking, however, they would not provide good views for vertical structures such as shelves. Moreover, in top-down sensor views, overlap is required for tracking hand-off between neighboring cameras \cite{tracking1, mittal2008general}, increasing the number of cameras  required specially if the roof is low.
On the other hand, frontal cameras could provide better coverage and clear views of the shelves, however, humans could be easily occluded by other people and obstacles.
The objective of this work is to find the pose of cameras within a bounded 3D environment to achieve user-defined coverage overlap under placement constraints with a limited number of cameras. The same formulation can be used for multi-robot system to compute their poses in a cooperative environmental survey application.
The contributions of the work are: 
1) The camera pose estimation is posed as a constrained optimization to minimize the squared difference between the user-defined desired coverage and the achieved coverage. The non-linear non-convex optimization is further reformulated as an equivalent mixed integer linear programming problem.
2) A framework to generate camera views and compute a coverage score in $3D$ while accounting for environmental occlusions. 

\section{Related Works}
\label{related}


The problem of sensor placement for coverage is known as the Art Gallery problem \cite{CHVATAL197539} \cite{10.5555/40599} which formulates the problem of finding the number and locations of stationary guards to fully cover a $2d$ space. Many variants of the problem have been proposed
such as the triangulation solution in \cite{FISK1978374}, randomized algorithms \cite{10.1145/378583.378674} and chromatic art gallery problem that computed the minimum number of colors required to color a guard set \cite{erickson2011many}. Also, there have been approaches that treat the placement problem in more realistic scenarios \cite{8820295}. 

Work has been done for the modeling of coverage of multi-sensor networks \cite{mavrinac2013modeling}. An automated camera layout approach satisfying task and floor plan-specific coverage requirements was proposed in \cite{ERDEM2006156}.
The work in \cite{placement_strategy} presented an approach for multi-camera placement in $3D$ volumetric spaces, also, camera models to compute good stereo-pair sensor placements for $3D$ reconstruction was presented in \cite{multiplecamera3d} where the authors provided recommendations for the angular baseline. 
An approach for modeling of sound sources in a $3D$ map utilizing probabilistic raycasting was proposed in \cite{sound}. 

There are different approaches regarding placement optimization works \cite{7424792}, \cite{10.1145/1178782.1178800}, \cite{doi:10.1111/mice.12385}, \cite{shm}. The approach in \cite{Shen2007AMS} proposed a definition of camera quality of view and a coverage term to rate camera configurations in $2D$. A probabilistic sensor model for coverage using gradient descent was proposed in \cite{gradient_descent}.
Also, \cite{suvsanj2020effective} proposed an approach for $2D$ and $3D$ sensor placement utilizing probabilistic coverage metric for optimization with genetic algorithms. 
The work in \cite{8758395} prioritized certain labeled regions named risk maps for sensor optimization in $2D$. Some other works center in sensor optimization for sensor overlap for target tracking \cite{6167509, 8335350, wirelessnetworkvisibility} and  efficient hand-offs \cite{5170013}.

The work presented in this paper uses real-world $3D$ environmental map representations that allows the computation of $6DoF$ camera poses constrained by the physical structures. The framework makes use of the camera model to project and raycast the camera pixels into the map; the result of the raycasting makes it possible to compute accurate coverage metrics. The raycasting framework can use different camera models such as pinhole or fish-eye.
Additionally, the proposed formulation allows the user to specify the desired coverage overlap at each location in the $3D$ space. Unlike the optimization formulations in earlier works 
which treat coverage at each location as a binary valued function (scored if desired coverage is attained and 0 otherwise), the proposed optimization treats coverage as an integer valued function, ensuring that the impact of coverage is more accurately captured, specially when the achieved coverage is lower than the desired coverage. 


\section{Problem Formulation}
\label{formulation}

The camera placement optimization problem is modeled 
as a combinatorial problem, wherein, given a set of possible camera poses $\bbG$, the objective is to select the subset $\bbS \subseteq \bbG$ under the 
coverage constraints. The cardinality of set $\bbG$ represents the total number of possible camera poses $N_g = |\bbG|$. The $i$-th camera's pose in set $\bbG$ is given as, $\bbG_i = \{x_i,y_i,z_i,\rho_i,\theta_i,\phi_i\}$, and represents the $6$ $DoF$ along which the camera's pose can be altered, the translations along the X,Y,Z axis and the rotations corresponding to yaw, pitch and roll angles. 

Let $\bbW$ be the set of all points in $\mathbb{R}^3$ that need to be observed by the camera. 
The $3D$ space/volume is quantized as voxels, where all voxels are cubes with a user-defined size. Thus, the i-th member of the set $\bbW$, ${\bbW}_i = \{x,y,z\}$ represents the position of the center of the voxel to be observed and $N_p = |{\bbW}|$ is the size of the set (see Section \ref{sec:implementation} for details).

\subsection{Multi-Camera Coverage} \label{sec:coverage}
A key factor to model the camera placement optimization is the overall visual coverage attained by the network of cameras. In realistic scenarios, different regions of the environment have different coverage requirements. In museum or store-like environments, showcases or shelves are required to be seen by at least one camera, while multi-camera overlap is required to track and localize people. Although, a two camera overlap is enough to find the $3D$ location of a point, to ensure redundancy and avoid people potentially occluding each other, an overlap of three or more cameras is required. 

To formalize the discussion around coverage, an important aspect is to model the set of voxels observed by each of the camera in $\bbG$. The view of the i-th camera ($\forall i \in \{1,...,N_g\}$) is thus given by the binary array $\bbV_{i:}$. The dimensions of $\bbV$ are $N_g \times N_p$ and the elements of $\bbV_{i:}$ with a value of 1 represent the voxels in the $3D$ space ${\bbW}$ which can be viewed by the i-th camera. 

The variable coverage requirements at the $j$-th voxel can then be expressed as minimizing the difference between the desired coverage, $\pmb\Gamma_j$, and the coverage score achieved by the optimization $\sum_{i=1}^{N_g} \bbV_{ij}\bbx_i$. The resulting cost across all voxels can thus be expressed as:

\begin{align}\label{eq:cost_1}
    &\sum_{j=1}^{N_p}(\operatorname{max}(\pmb\Gamma_j - \sum_{i=1}^{N_g} \bbV_{ij}\bbx_i~,~ 0))^2 \nonumber\\
           &\text{s.t.} ~  \sum_{i=1}^{N_g} \bbx_i \le {N_s}
\end{align}
where, the array indicating the required/target coverage, $\pmb{\Gamma}$, is of the same size as ${\bbW}$ ($N_p\times 1$). The array, $\pmb{\Gamma}$, has integer values indicating the required coverage at each location, thus, a value of 3 would indicate that three camera overlap is required at a specific location. $\bbx \in  \mathbb{B}^{N_g}$, is the the camera selection variable and indicates whether a selected camera is present at any of the $N_g$ location. Thus, for all the cameras in $\bbS$ the variable $\bbx_i = 1$, and is $0$ else where.  The term $\operatorname{max}(\pmb\Gamma_j - \sum_{i=1}^{N_g} \bbV_{ij}\bbx_i, 0)$ ensures that the cost/error function is affected only if the achieved coverage is lower than the target coverage and accounts for 0 error otherwise. 
The constraint in \eqref{eq:cost_1} limits the maximum number of cameras that can be selected as part of the optimization to a specified number $N_s$, which depicts the available budget of cameras.

Previous work formulations \cite{10.1145/1178782.1178800,Zhao2008,zhao2009optimal} only considered the voxels that satisfy the required overlap coverage to contribute towards the optimization cost function ignoring the voxels with lower camera overlaps.
This prioritizes camera placements that satisfies the three camera voxel overlap, at the cost of leaving other voxels unobserved. 
Alternatively, the formulation in this work takes into consideration all the voxels irrespective of the camera overlap values, and utilizes them in the cost function. Also, the squared error in \eqref{eq:cost_1} prioritizes covering voxels with lower coverage (corresponding to quadratically higher cost/error) as compared to increasing the coverage in voxels with coverage close to the desired levels. Thus, the overall placement optimization results in a coverage where more voxels are observed and has an overall result closer to the desired coverage. 




\subsection{Angle of View} \label{sec:angle_of_view}
The coverage related metrics of Section \ref{sec:coverage} maximize the number of voxels covered by the cameras, but the angle at which these voxels are observed could be steep, thus making the views inappropriate for certain practical applications. 
For some applications where it is important to observe a region of interest such as a showcase or shelf, it is important that the  angle of view/incidence, (i.e. angle between a raycasted pixel of the camera and the normal to the surface of interest) be low. 
Instead of introducing these constraints as part of the optimization problem, they can be effectively imposed during the construction of the camera's view $\bbV_i, \forall i \in \{1,...,N_g\}$. All the voxels for which the angle of incidence constraints are not satisfied, can be removed from the $\bbV_i$. Such a strategy essentially reduces the computational complexity of the overall optimization.

\subsection{Location Constraints}
To efficiently pose the constraint regarding having only one camera at each location, we form the matrix $\tilde{\bbL}$ of size $N_l \times N_g$, where $N_l$ is the number of unique location in the set $\bbG$ (as the set $\bbG$ may have elements with the same location but different orientations). For the i-th row, $\tilde{\bbL}_{i:}$, the dimensions having a value of 1 indicate that the corresponding elements in the set $\bbG$ have the same location in the $3D$ space.  All other dimensions of $\tilde{\bbL}_{i:}$ are given a value of 0. The corresponding constraint can be expressed as,
\begin{align}\label{eq:location_constraint}
    \sum_{i=1}^{N_g} \tilde{\bbL}_{ij}\bbx_j \le 1~~~ \forall j \in \{1,...,N_l\}.
\end{align}

The overall optimization cost is then expressed as:

\begin{align}\label{eq:opt_cost}
   &\arg\min_{\bbx}  \sum_{j=1}^{N_p}(\operatorname{max}(\pmb\Gamma_j - \sum_{i=1}^{N_g} \bbV_{ij}\bbx_i~,~ 0))^2 \nonumber\\
   & \text{s.t.} ~  \sum_{i=1}^{N_g} \bbx_i \le {N_s}, ~\sum_{i=1}^{N_g} \tilde{\bbL}_{ij}\bbx_j \le 1~~~ \forall j \in \{1,...,N_l\}.
\end{align}

\section{Optimization}
\label{sec:optimization}

The combinatorial optimization  problem in \eqref{eq:opt_cost} is non-linear. This is essentially due to the $\operatorname{max}(\cdot)$ and the squared operators in the cost function. Next, we linearize the cost function 
while ensuring that the reformulated cost is equivalent to the original in \eqref{eq:opt_cost}. 
Using the epigraph approach we can linearize the cost and push the non-linearities to the constraints, thus we have:

\begin{align}
    &\arg\min \sum_{j=1}^{N_p}(\operatorname{max}(\pmb\Gamma_j - \sum_{i=1}^{N_g} \bbV_{ij}\bbx_i,0))^2 <=>\nonumber\\
    &  \arg\min \sum_{j=1}^{N_p} (\bbQ_j)^2 \nonumber \\
    & \text{s.t.} ~~~ \bbQ_j \ge \operatorname{max}(\pmb\Gamma_j - \sum_{i=1}^{N_g} \bbV_{ij}\bbx_i,0).
\end{align}

Since the desired coverage variable $\pmb\Gamma_j$ and the achieved coverage $ \sum_{i=1}^{N_g} \bbV_{ij}\bbx_i$ are both integer values, we can further linearize the cost using an incremental piecewise linear model \cite{piecewise}.



Considering the case, where the maximum required coverage at any location is 3, thus $\pmb\Gamma_j = 3, \forall j$, the piecewise linear formulation can be expressed as: 
\begin{align}\label{eq:pwln_2}
    & \arg\min \sum_{j=1}^{N_p} 0f^0_j+1f^1_j+2f^2_j+3f^3_j \nonumber \\
    &\text{s.t.} ~~~ \bbQ_j - (\pmb\Gamma_j - \sum_{i=1}^{N_g} \bbV_{ij}\bbx_i) \le 0f^0_j+f^1_j+f^2_j+f^3_j ,\nonumber\\
    & lb~  pl^0_j < f^0_j <= 0.5 ~  pl^0_j\nonumber\\
    & 0.5~ pl^1_j< f^1_j <= 1.5~  pl^1_j\nonumber\\
    & 1.5~  pl^2_j< f^2_j <= 2.5~  pl^2_j\nonumber\\
    & 2.5~  pl^3_j< f^3_j <= \pmb\Gamma_j~  pl^3_j ~~~~~ \nonumber\\
    & pl^0_j +pl^1_j+pl^2_j+pl^3_j = 1 ~~~ \forall j \in \{1, ..., Np\} \nonumber\\
    & \bbx \in \mathbb{B}^{N_g}, \{pl^0,pl^1,pl^2,pl^3\}\in \mathbb{B}^{N_p},\nonumber\\
    &\{f^0,f^1,f^2,f^3\}\in \mathbb{I}^{N_p}
\end{align}

In the above model since $\bbQ_j$ can only take integer values between $[0,3]$, we split $\bbQ_j$ in 4 linear pieces. Thus, $pl^0_j,pl^1_j,pl^2_j,pl^3_j$ are the 4 corresponding piece-wise linear variables which can take a value of $0$ or $1$. The constraint $pl^0_j +pl^1_j+pl^2_j+pl^3_j = 1$ ensures that only one of the pieces is operational at any given time in the optimization. $\{0, f^1_j =1, f^2_j = 2,f^3_j=3$\} are the possible values that $\bbQ_j$ can achieve corresponding to the 4 linear pieces. $lb$ is the lower bound on $\pmb\Gamma_j - \sum_{i=1}^{N_g} \bbV_{ij}\bbx_i$. The 4 piecewise linear pieces ensure that the function $\bbQ_j - (\pmb\Gamma_j - \sum_{i=1}^{N_g} \bbV_{ij}\bbx_i)$ can only take values in the set $\{0,1,2,3\}$. Thus, the overall, mixed integer linear programming problem can be expressed as the following constrained optimization:


\begin{align}\label{eq:cov_max}
    & \arg\min \sum_{j=1}^{N_p} 0f^0_j+1f^1_j+2f^2_j+3f^3_j \nonumber \\
    &\text{s.t.} ~~~ \bbQ_j - (\pmb\Gamma_j - \sum_{i=1}^{N_g} \bbV_{ij}\bbx_i) \le 0f^0_j+f^1_j+f^2_j+f^3_j ,\nonumber\\
    & lb~ pl^0_j < f^0_j <= 0.5~ pl^0_j\nonumber\\
    & 0.5~ pl^1_j< f^1_j <= 1.5~ pl^1_j\nonumber\\
    & 1.5~ pl^2_j< f^2_j <= 2.5~ pl^2_j\nonumber\\
    & 2.5~ pl^3_j< f^3_j <= \pmb\Gamma_j~ pl^3_j  \nonumber\\
    & pl^0_j +pl^1_j+pl^2_j+pl^3_j = 1 ~~~~~\forall j \in \{1, ..., Np\}, \nonumber\\
    & \bbx \in \mathbb{B}^{N_g}, \{pl^0,pl^1,pl^2,pl^3\}\in \mathbb{B}^{N_p},\nonumber\\
    &\{f^0,f^1,f^2,f^3\}\in \mathbb{I}^{N_p},\nonumber \\
    &  \sum_{i=1}^{N_g} \bbx_i \le {N_s}~, ~ \sum_{i=1}^{N_g} \tilde{\bbL}_{ij}\bbx_j \le 1~~~ \forall j \in \{1,...,N_l\}
\end{align}






For solving the mixed integer optimization problem, a branch and bound method with cutting planes was utilized \cite{Nemhauser1988}. For the implementation we used the open source Python-MIP library \cite{PyMIP}.


\section{Camera View Computation and Coverage Scoring}
\label{sec:implementation}

\begin{figure}[tb]
 \centering
 \includegraphics[width=3in, angle=0]{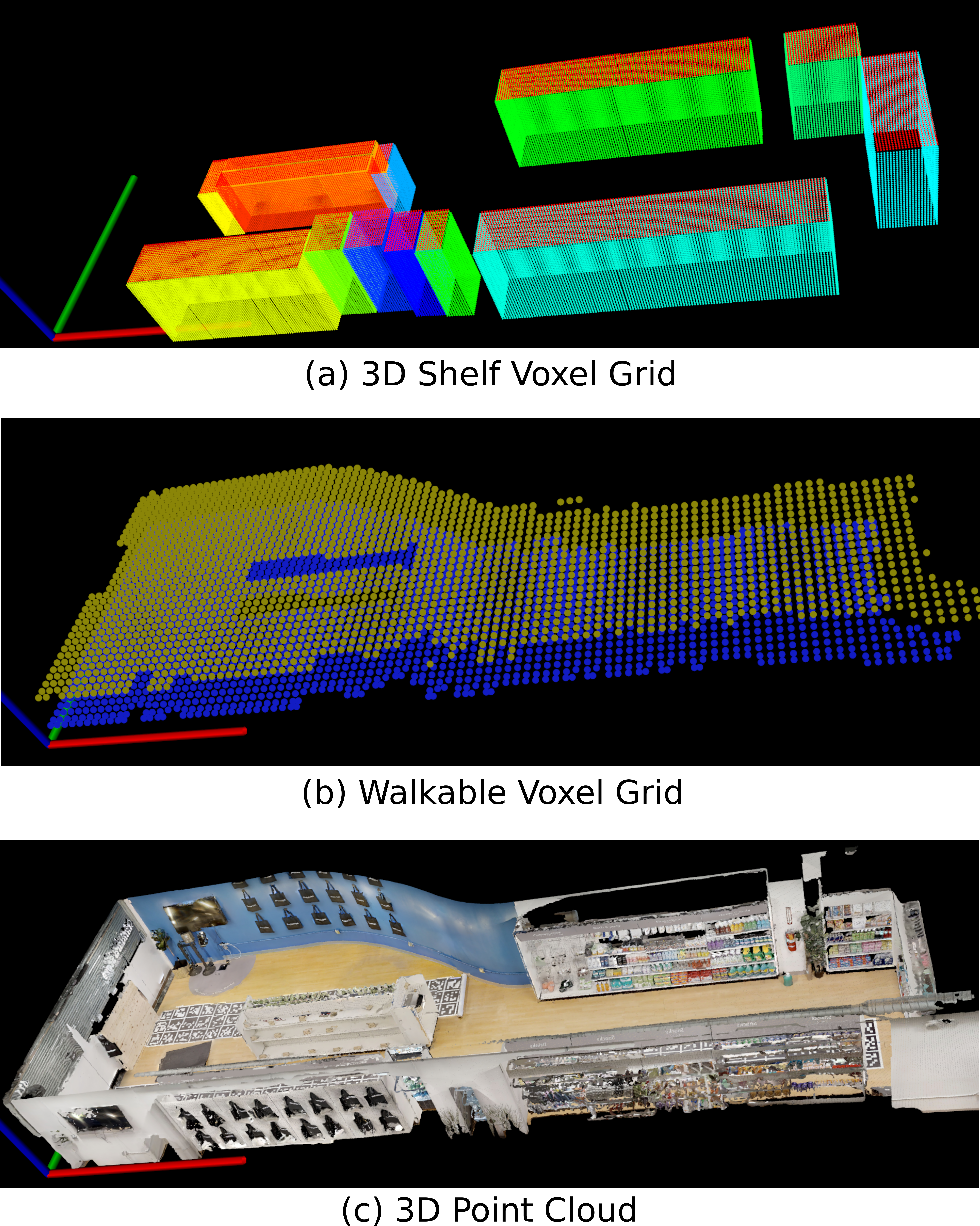}
   \caption{Environmental representation $\bbW$. In the bottom, the colored $3D$ map as a point cloud. In the middle, two free space planes represented as voxel maps at heights of $0.5m$ and $1.5m$ which represent the walkable areas. In the top the shelves represented as voxel grid planes.} 
\label{fig:cloud}
\end{figure}

The evaluation of the proposed against existing camera placement approaches was done in two real convenience stores environments.
We used $3D$ dense point cloud maps as environmental geometric models $\bbW$. The point clouds were downsampled to a tree-based voxel data structure which allows to perform raycasting from a camera to the environment. A colored voxel map of a store with resolution of $0.01$ m is shown in the bottom of Fig. \ref{fig:cloud}. 
In order to compute environmental coverage in the walkable empty space, a free-space voxel grid was prepared. Two traversal planes of the empty space grid at a height of $0.5$ m and $1.5$ m are shown in the middle of Fig. \ref{fig:cloud}.
To compute coverage for the shelves which are the regions of interest, the shelves in the map were labeled and modeled as  $3D$ boxes as shown on top of Fig. \ref{fig:cloud} where each shelf has a different color (with the top of shelves in red). For all our experiments we have used a voxel size of $0.25$ m.

\begin{figure}[tb]
 \centering
 \includegraphics[width=3in, angle=0]{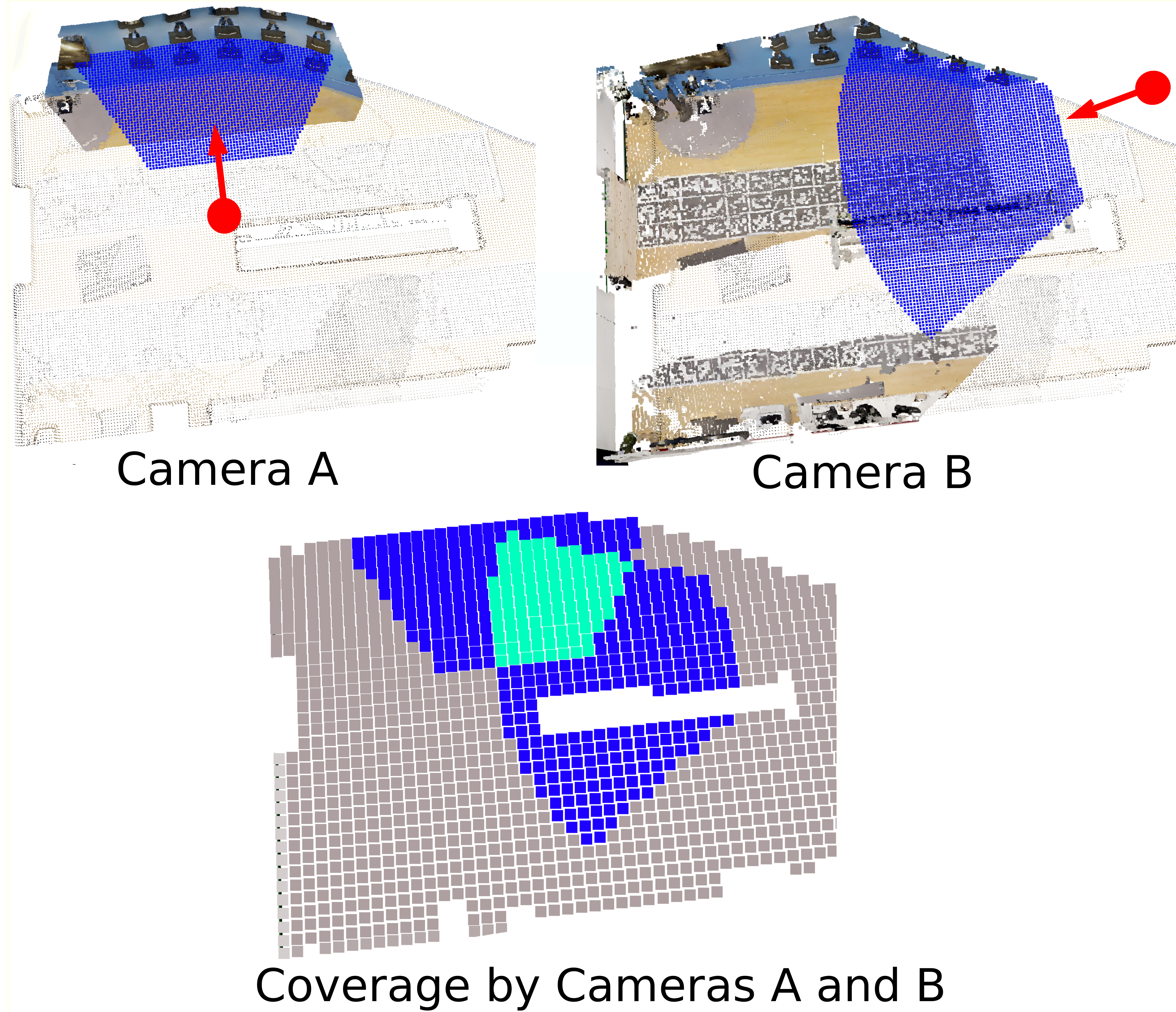}
   \caption{Illustration of the coverage of two cameras A and B. The top section shows two cameras A and B in red. The $3D$ raycasting is colored in the map and the raycast projection on a grid at a height of $1.5$ $m$ is shown in blue. The bottom shows the coverage of both cameras, in blue the voxels covered by a single camera and in cyan the voxels covered by both cameras.} 
\label{fig:coverage}
\end{figure}

For the experimentation presented in this work we used a camera pinhole model with a field of view of $71^o \times 36^o$ and a resolution of $1780\times720$ (horizontal x vertical). 
Each $i^{th}$ pixel $i_i,j_i$ of the camera screen are projected through the camera pinhole model to the voxel map through a raycasting function obtaining the endpoint $x_i, y_i, z_i$ in world coordinates. The useful maximum distance of a camera was set to $5$ m.
Fig. \ref{fig:coverage} shows an example of two overlapping cameras being raycasted from their pose $\bbG_i = \{x_i,y_i,z_i,\rho_i,\theta_i,\phi_i\}$ 
towards the environment ${\bf W}$. The top of Fig. \ref{fig:coverage} shows the colored raycasted voxels of cameras A and B in color, the resulting raycasted voxels in the walkable voxel grid are shown in blue. 
The bottom of Fig. \ref{fig:coverage} shows the sum of the covered voxels. 
Blue voxels show $1$ camera coverage and cyan voxels show $2$ camera coverage. The same process is repeated to raycast and build the view matrix $\bbV_{i:}$ for each camera in $\bbG$. This is the basis to compute the coverage score of $\bbG$ in an environment.

\begin{figure}[!t]
  \centering
  \includegraphics[width=70mm]{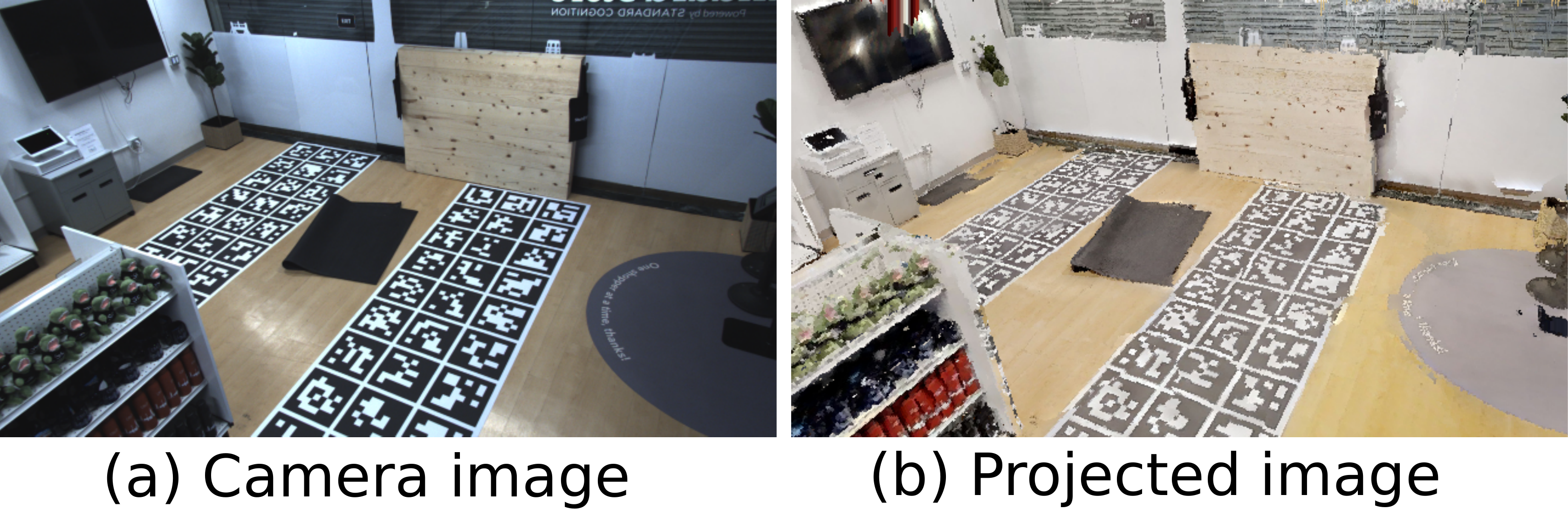}
	\caption{Example of a camera image and its projection computed utilizing the a pinhole camera model and its calibration parameters.}
  \label{fig:projection}
\end{figure}

An illustration of the raycasting framework used to compute camera coverage is shown in Fig. \ref{fig:projection}. Fig. \ref{fig:projection}(a) shows a real camera image and Fig. \ref{fig:projection}(b) shows the raycasting generated image which illustrates the realistic projection of the pinhole model through its extrinsic parameters.
The raycasting could be computed with different camera lens models such as fisheye.


\section{Experimental Procedure and Results}
\label{results}
To show the efficacy of our proposed approach we evaluate the performance of our camera placement optimization approach in two different stores. 1) The first one corresponds to a convenience store space owned by Standard Cognition located in San-Francisco with an area of $116.4$ m$^2$. 2) The second is a convenience store operated by a large retail chain which for confidentiality, we name 'Store A'. It has an area of  $180.7$ m$^2$.

We compare the performance of the proposed mixed integer formulation ({Proposed MIP}) with two other methods. 1) The mixed integer optimization formulation proposed by Zhao et. al. \cite{Zhao2008}, 2) The greedy formulation proposed by Horster et. al. \cite{10.1145/1178782.1178800}. We also report the performance of the proposed optimization formulation if solved as a greedy optimization using the strategy explained in \cite{10.1145/1178782.1178800} (Proposed Greedy)

The existing methods in  \cite{Zhao2008,10.1145/1178782.1178800} have been proposed for problems similar to environmental monitoring and not specifically for store like environments with specific regions of interest (like shelves) having specialized coverage requirements. The results have been computed and presented only for the coverage of the $3D$ walkable voxel grid, as shown in Fig. \ref{fig:cloud}(b). 

To evaluate the performance of the different methods towards their usability in real settings, we compare them across two different criterion. First, to capture the coverage perspective, we define the coverage optimality gap as the normalized difference between achieved coverage and the specified desired target coverage across all $N_g$ voxels. The smaller the value for the coverage optimality gap, the better the performance. The coverage optimality gap is defined as:  

\begin{align} \label{cov_opt_gap}
    \frac{\sum_{j=1}^{N_p}(\operatorname{max}(\pmb\Gamma_j - \sum_{i=1}^{N_g} \bbV_{ij}\bbx_i~,~ 0))^2}{\sum_{j=1}^{N_p} \pmb\Gamma_j^2} 
\end{align}

To capture the usability of the achieved camera placement configuration from a triangulation perspective, we evaluate the percentage of voxels that cannot be triangulated, which we define as 
any voxel covered by less than two cameras. 


For each experiment, to build the set, $\bbG$, we consider a grid of spacing $1$ m $\times$ $1$ m across the roof of the store. It is assumed that the cameras can be placed at the corners of this grid.  At each location, the, yaw $\rho$, can take values in the range $[0-360]$ degrees, in steps of 30 degrees, and the pitch, $\theta$, can take values of $\{30, 45, 60\}$. The roll, $\phi$, is kept fixed at $0$ degrees. Some views which may be completely blocked by the walls or roof fixtures, are removed from the set before optimization.
\subsection{Standard Store Camera Placement} \label{sec:stcg}
For the comparison of the four placement methods in the standard store environment, we consider a set of $|\bbG| = N_g = 3696$ possible camera poses. For a fair comparison between the methods, we set a maximum allowable time of 6000 seconds in a unicore setting for the MIP based optimization methods (Zhao et. al. \cite{Zhao2008} and the proposed optimization). In the multi-core setting (with 12 cores) in which we run the MIP optimization, this would result in an effective maximum time of approximately ($6000/12 \approx  500$) seconds. If the method does not reach the optimal solution within the defined time, the best known solution that was found within the maximum allowable time is used.

\begin{figure}[!t]
  \centering
  \includegraphics[width=3.5in,height= 2in]{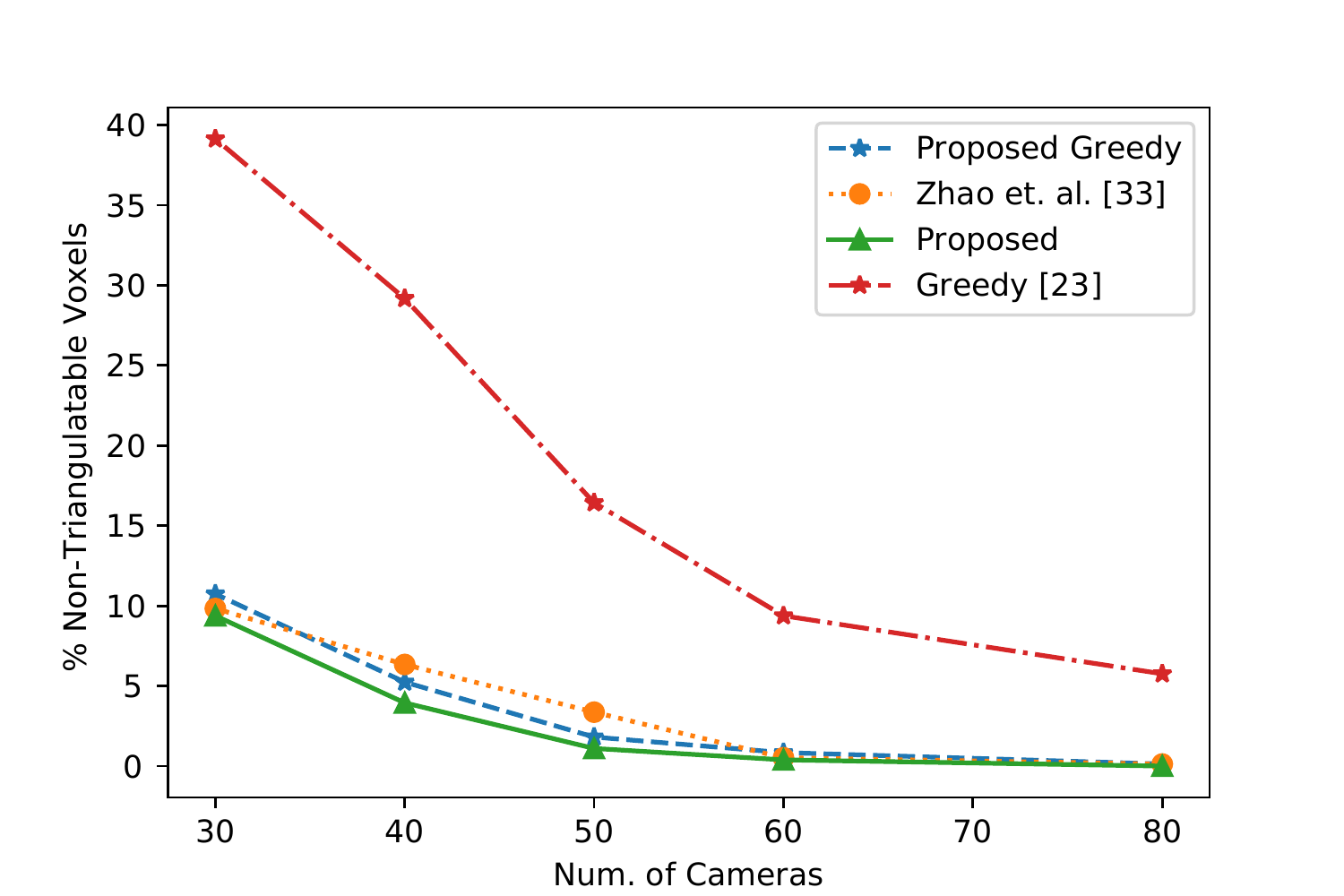}
	\caption{Percentage of non-triangulatable voxels in the Standard store using camera placements attained from different methods. A voxel is said to be non-triangulatable if it is seen by less than 2 cameras.}
  \label{fig:stcg}
\end{figure}

Table \ref{table:stcg} presents the values of coverage optimality gap (as defined in \eqref{cov_opt_gap}) for different values of $N_s$. As can be inferred from the table, the coverage achieved by the proposed formulation out performs the existing methods in \cite{Zhao2008,10.1145/1178782.1178800}. A similar trend can be observed in the comparison in Fig. \ref{fig:stcg} which shows the \% of voxels having insufficient coverage for triangulation. The proposed formulation provides camera placements with better performance in overall coverage and ensures that more voxels are observed by greater than 2 cameras. 

\begin{table}[!t]
\centering
\caption{Coverage Optimality Gap for Standard Store Data}
\begin{tabular}{|l|cccccc|}
\hline
\textbf{} & \multicolumn{6}{c|}{\textbf{Number of Cameras ($N_s$)}}   \\
  & \textbf{80} & \textbf{60}& \textbf{50}&\textbf{40}&\textbf{30} & \textbf{20} \\
\hline
 \textbf{Zhao et. al. \cite{Zhao2008}} & 0.0 &0.01& 0.03& 0.04& 0.09& 0.23 \\
 \textbf{Greedy \cite{10.1145/1178782.1178800}} & 0.04 &0.07 & 0.13 &0.26 & 0.34& 0.47 \\
 \textbf{Proposed Greedy} & \bf{0} & 0.01& 0.02.     & \bf{0.04} & \bf{0.08} &\bf{0.16}\\
  \textbf{Proposed MIP}     & \bf{0}& \bf{0}& \bf{0.01}& \bf{0.04}& \bf{0.08}& \bf{0.16} \\

\hline
\end{tabular}
\label{table:stcg}
\end{table}

\subsection{Store A Camera Placement} \label{sec:circlek}
\begin{figure}[!t]
  \centering
  \includegraphics[width=3.5in,height= 2in]{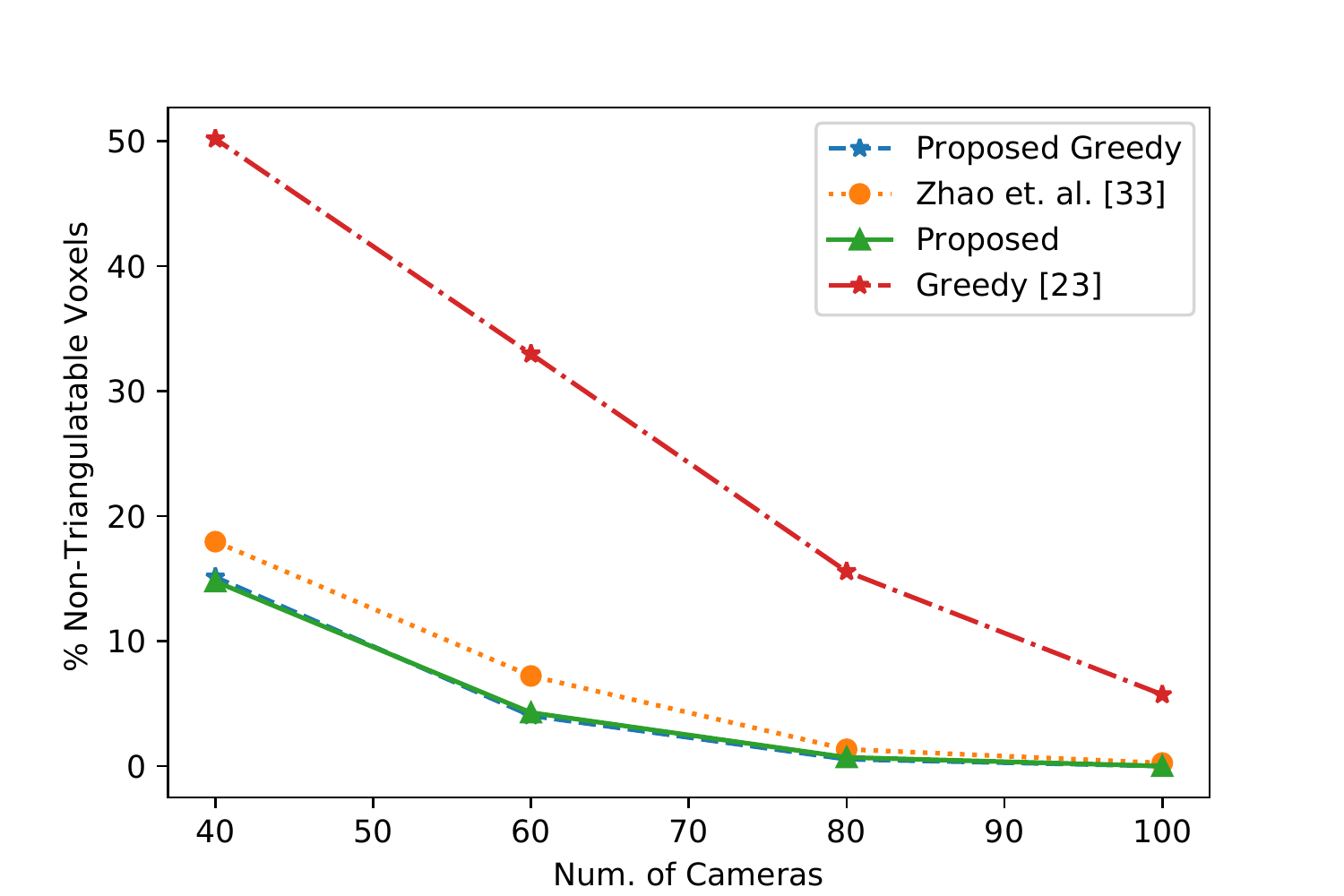}
	\caption{Percentage of non-triangulatable voxels in the store A using camera placements attained from different methods. A voxel is said to be non-triangulatable if it is seen by less than 2 cameras.}
  \label{fig:Circle_K}
\end{figure}

For the comparison of the four placement methods in the second environment, we consider a set of $|\bbG| = N_g = 3312$ possible camera poses. As seen Table \ref{table:Circle_K}, all the methods reach close to the required coverage with $N_s = 100$ cameras. For fewer cameras, the coverage optimality gap obtained with the proposed formulation is lower than all the other methods. The performance improvement is particularly evident in the case with $N_s = 20,~40$. Similar performance improvement can be seen in Fig. \ref{fig:Circle_K}, where the proposed formulation is improves the percentage of triangulatable points.

Interestingly, in both the experimental setups, the difference in performance of the proposed formulation with a greedy solver and with the MIP solver is minimal in the coverage metric. While the MIP optimization seems to provide a marginal improvement in terms of the \% of triangulatable voxels in the first case of Standard Store (Fig. \ref{fig:stcg}) but a similar performance in the second case (Fig. \ref{fig:Circle_K}). Thus, in a setting where the placement optimization needs to be evaluated under limited time constraints, the proposed formulation, can be solved with a greedy solver, whereas in the scenarios where no time constraints apply, the MIP solver computes placements with better results.

Our formulation can include the coverage of regions of interest (e.g., shelves of Fig. \ref{fig:cloud}(a)) by assigning the required $\pmb \Gamma_j$ for all $j$ corresponding to the shelf voxels and utilizing the angle of incidence constraints as specified in Section \ref{sec:angle_of_view}. As previous works did not consider this feature, an experimental evaluation was not performed. However, this is a relevant feature for real applications, as it allows to specify the desired camera coverage for different locations in the environment. 

For all evaluations in both environments the camera sets $|\bbG|$ were built with the approach of Section \ref{sec:implementation}, which allows to compute accurate metrics not available in the original works \cite{10.1145/1178782.1178800, Zhao2008}. It is left for future work to formalize the cost function to compute the minimum number of cameras under user specified constraints. 

\begin{table}[!t]
\centering
\caption{Coverage Optimality Gap for Store A Data}
\begin{tabular}{|l|ccccc|}
\hline
\textbf{} & \multicolumn{5}{c|}{\textbf{Number of Cameras ($N_s$)}}   \\
 & \textbf{100} & \textbf{80} & \textbf{60}& \textbf{40} & \textbf{20} \\
\hline

 \textbf{Zhao et. al. \cite{Zhao2008}} & 0 & 0.01 & 0.06 & 0.13 & 0.41 \\
 \textbf{Greedy \cite{10.1145/1178782.1178800}} & 0.04 & 0.12 & 0.30 & 0.48  &0.64 \\
 \textbf{Proposed Greedy} & \bf{0} & \bf{0.01} & \bf{0.04}  & \bf{0.12} & 0.35 \\
   \textbf{Proposed MIP} & \bf{0}    & \bf{0.01} & \bf{0.04} & \bf{0.12} & \bf{0.34} \\

\hline
\end{tabular}
\label{table:Circle_K}
\end{table}

\section{Conclusions}
\label{sec:conclusions}

This work presented a novel end to end camera $6$ $DoF$ placement framework for real environments. 
Given a maximum number of cameras, the camera placement is modeled as mixed integer linear programming problem that selects a subset of camera poses that minimizes the squared difference between the desired and the achieved coverage, while satisfying the posed constraints. 
Additionally, a raycasting framework which allows to score a camera set coverage while taking into consideration environmental occlusions was proposed and utilized to evaluate the performance of the proposed and existing camera placement methods. 
Experimental results show that the performance of the proposed approach is higher than previous camera placement schemes in terms of coverage and overlap for triangulation, especially when the total number of cameras is low.

\bibliographystyle{IEEEtran} 
\bibliography{IEEEabrv,references}

\begin{thebibliography}{10}
\providecommand{\url}[1]{#1}
\def\UrlFont{\rmfamily}
\providecommand{\newblock}{\relax}
\providecommand{\bibinfo}[2]{#2}
\providecommand\BIBentrySTDinterwordspacing{\spaceskip=0pt\relax}
\providecommand\BIBentryALTinterwordstretchfactor{4}
\providecommand\BIBentryALTinterwordspacing{\spaceskip=\fontdimen2\font plus
\BIBentryALTinterwordstretchfactor\fontdimen3\font minus
  \fontdimen4\font\relax}
\providecommand\BIBforeignlanguage[2]{{%
\expandafter\ifx\csname l@#1\endcsname\relax
\typeout{** WARNING: IEEEtran.bst: No hyphenation pattern has been}%
\typeout{** loaded for the language `#1'. Using the pattern for}%
\typeout{** the default language instead.}%
\else
\language=\csname l@#1\endcsname
\fi
#2}}

\bibitem{wang2011coverage}
B.~Wang, ``Coverage problems in sensor networks: A survey,'' \emph{ACM
  Computing Surveys (CSUR)}, vol.~43, no.~4, pp. 1--53, 2011.

\bibitem{Shen2007AMS}
C.~Shen, C.~Zhang, and S.~Fels, ``A multi-camera surveillance system that
  estimates quality-of-view measurement,'' \emph{2007 IEEE International
  Conference on Image Processing}, vol.~3, pp. III -- 193--III -- 196, 2007.

\bibitem{brvsvcic2013person}
D.~Br{\v{s}}{\v{c}}i{\'c}, T.~Kanda, T.~Ikeda, and T.~Miyashita, ``Person
  tracking in large public spaces using 3-d range sensors,'' \emph{IEEE
  Transactions on Human-Machine Systems}, vol.~43, no.~6, pp. 522--534, 2013.

\bibitem{urrutia2000art}
J.~Urrutia, ``Art gallery and illumination problems,'' in \emph{Handbook of
  computational geometry}.\hskip 1em plus 0.5em minus 0.4em\relax Elsevier,
  2000, pp. 973--1027.

\bibitem{amazongo}
\BIBentryALTinterwordspacing
Amazon. (2021) Amazon go description. [Online]. Available:
  \url{https://www.amazon.com/b?node=16008589011}
\BIBentrySTDinterwordspacing

\bibitem{zippin}
\BIBentryALTinterwordspacing
Zippin. (2021) Zippin description. [Online]. Available:
  \url{https://www.getzippin.com/}
\BIBentrySTDinterwordspacing

\bibitem{liu2020grab}
X.~Liu, Y.~Jiang, K.-H. Kim, and R.~Govindan, ``Grab: Fast and accurate sensor
  processing for cashier-free shopping,'' 2020.

\bibitem{ruiz2019aim3s}
C.~Ruiz, J.~Falcao, S.~Pan, H.~Y. Noh, and P.~Zhang, ``Aim3s: Autonomous
  inventory monitoring through multi-modal sensing for cashier-less convenience
  stores,'' in \emph{Proceedings of the 6th ACM International Conference on
  Systems for Energy-Efficient Buildings, Cities, and Transportation}.\hskip
  1em plus 0.5em minus 0.4em\relax ACM, 2019, pp. 135--144.

\bibitem{tracking1}
Y.~Yao, C.-H. Chen, B.~Abidi, D.~Page, A.~Koschan, and M.~Abidi, ``Sensor
  planning for automated and persistent object tracking with multiple
  cameras,'' \emph{2012 IEEE Conference on Computer Vision and Pattern
  Recognition}, vol.~0, pp. 1--8, 06 2008.

\bibitem{mittal2008general}
A.~Mittal and L.~S. Davis, ``A general method for sensor planning in
  multi-sensor systems: Extension to random occlusion,'' \emph{International
  Journal of Computer Vision}, vol.~76, no.~1, pp. 31--52, 2008.

\bibitem{CHVATAL197539}
\BIBentryALTinterwordspacing
V.~Chvátal, ``A combinatorial theorem in plane geometry,'' \emph{Journal of
  Combinatorial Theory, Series B}, vol.~18, no.~1, pp. 39 -- 41, 1975.
  [Online]. Available:
  \url{http://www.sciencedirect.com/science/article/pii/0095895675900611}
\BIBentrySTDinterwordspacing

\bibitem{10.5555/40599}
J.~O'Rourke, \emph{Art Gallery Theorems and Algorithms}.\hskip 1em plus 0.5em
  minus 0.4em\relax USA: Oxford University Press, Inc., 1987.

\bibitem{FISK1978374}
\BIBentryALTinterwordspacing
S.~Fisk, ``A short proof of chvátal's watchman theorem,'' \emph{Journal of
  Combinatorial Theory, Series B}, vol.~24, no.~3, p. 374, 1978. [Online].
  Available:
  \url{https://www.sciencedirect.com/science/article/pii/009589567890059X}
\BIBentrySTDinterwordspacing

\bibitem{10.1145/378583.378674}
\BIBentryALTinterwordspacing
H.~Gonz\'{a}lez-Banos, ``A randomized art-gallery algorithm for sensor
  placement,'' in \emph{Proceedings of the Seventeenth Annual Symposium on
  Computational Geometry}, ser. SCG '01.\hskip 1em plus 0.5em minus 0.4em\relax
  New York, NY, USA: Association for Computing Machinery, 2001, p. 232–240.
  [Online]. Available: \url{https://doi.org/10.1145/378583.378674}
\BIBentrySTDinterwordspacing

\bibitem{erickson2011many}
L.~H. Erickson and S.~M. LaValle, ``How many landmark colors are needed to
  avoid confusion in a polygon?'' in \emph{2011 IEEE International Conference
  on Robotics and Automation}.\hskip 1em plus 0.5em minus 0.4em\relax IEEE,
  2011, pp. 2302--2307.

\bibitem{8820295}
J.~{Kritter}, M.~{Brévilliers}, J.~{Lepagnot}, and L.~{Idoumghar}, ``On the
  real-world applicability of state-of-the-art algorithms for the optimal
  camera placement problem,'' in \emph{2019 6th International Conference on
  Control, Decision and Information Technologies (CoDIT)}, 2019, pp.
  1103--1108.

\bibitem{mavrinac2013modeling}
A.~Mavrinac and X.~Chen, ``Modeling coverage in camera networks: A survey,''
  \emph{International journal of computer vision}, vol. 101, no.~1, pp.
  205--226, 2013.

\bibitem{ERDEM2006156}
\BIBentryALTinterwordspacing
U.~M. Erdem and S.~Sclaroff, ``Automated camera layout to satisfy task-specific
  and floor plan-specific coverage requirements,'' \emph{Computer Vision and
  Image Understanding}, vol. 103, no.~3, pp. 156 -- 169, 2006, special issue on
  Omnidirectional Vision and Camera Networks. [Online]. Available:
  \url{http://www.sciencedirect.com/science/article/pii/S1077314206000671}
\BIBentrySTDinterwordspacing

\bibitem{placement_strategy}
\BIBentryALTinterwordspacing
I.~Fedorov, N.~Lawal, M.~O'Nils, and H.~Alqaysi, ``Placement strategy of
  multi-camera volumetric surveillance system for activities monitoring,'' in
  \emph{Proceedings of the 11th International Conference on Distributed Smart
  Cameras}, ser. ICDSC 2017.\hskip 1em plus 0.5em minus 0.4em\relax New York,
  NY, USA: Association for Computing Machinery, 2017, p. 113–118. [Online].
  Available: \url{https://doi.org/10.1145/3131885.3131911}
\BIBentrySTDinterwordspacing

\bibitem{multiplecamera3d}
J.~Starck, A.~Maki, S.~Nobuhara, A.~Hilton, and T.~Matsuyama, ``The
  multiple-camera 3-d production studio,'' \emph{Circuits and Systems for Video
  Technology, IEEE Transactions on}, vol.~19, pp. 856 -- 869, 07 2009.

\bibitem{sound}
J.~Even, J.~Furrer, Y.~Morales, C.~T. Ishi, and N.~Hagita, ``Probabilistic 3-d
  mapping of sound-emitting structures based on acoustic ray casting,''
  \emph{IEEE Transactions on Robotics}, vol.~33, no.~2, pp. 333--345, April
  2017.

\bibitem{7424792}
G.~{Zhang}, B.~{Dong}, and J.~{Zheng}, ``Visual sensor placement and
  orientation optimization for surveillance systems,'' in \emph{2015 10th
  International Conference on Broadband and Wireless Computing, Communication
  and Applications (BWCCA)}, 2015, pp. 1--5.

\bibitem{10.1145/1178782.1178800}
\BIBentryALTinterwordspacing
E.~H\"{o}rster and R.~Lienhart, ``On the optimal placement of multiple visual
  sensors,'' in \emph{Proceedings of the 4th ACM International Workshop on
  Video Surveillance and Sensor Networks}, ser. VSSN '06.\hskip 1em plus 0.5em
  minus 0.4em\relax New York, NY, USA: Association for Computing Machinery,
  2006, p. 111–120. [Online]. Available:
  \url{https://doi.org/10.1145/1178782.1178800}
\BIBentrySTDinterwordspacing

\bibitem{doi:10.1111/mice.12385}
\BIBentryALTinterwordspacing
X.~Yang, H.~Li, T.~Huang, X.~Zhai, F.~Wang, and C.~Wang, ``Computer-aided
  optimization of surveillance cameras placement on construction sites,''
  \emph{Computer-Aided Civil and Infrastructure Engineering}, vol.~33, no.~12,
  pp. 1110--1126, 2018. [Online]. Available:
  \url{https://onlinelibrary.wiley.com/doi/abs/10.1111/mice.12385}
\BIBentrySTDinterwordspacing

\bibitem{shm}
G.~Gomes, F.~Almeida, P.~Alexandrino, S.~Cunha~Jr, B.~Silva~de Sousa, and
  A.~Ancelotti, ``A multiobjective sensor placement optimization for shm
  systems considering fisher information matrix and mode shape interpolation,''
  \emph{Engineering with Computers}, p. 519–535, 05 2018.

\bibitem{gradient_descent}
V.~Akbarzadeh, J.-C. Lévesque, C.~Gagné, and M.~Parizeau, ``Efficient sensor
  placement optimization using gradient descent and probabilistic coverage,''
  \emph{Sensors (Basel, Switzerland)}, vol.~14, pp. 15\,525--52, 08 2014.

\bibitem{suvsanj2020effective}
D.~Su{\v{s}}anj, D.~Pin{\v{c}}i{\'c}, and K.~Lenac, ``Effective area coverage
  of 2d and 3d environments with directional and isotropic sensors,''
  \emph{IEEE Access}, vol.~8, pp. 185\,595--185\,608, 2020.

\bibitem{8758395}
A.~A. {Altahir}, V.~S. {Asirvadam}, N.~H.~B. {Hamid}, P.~{Sebastian}, M.~A.
  {Hassan}, N.~B. {Saad}, R.~{Ibrahim}, and S.~C. {Dass}, ``Visual sensor
  placement based on risk maps,'' \emph{IEEE Transactions on Instrumentation
  and Measurement}, vol.~69, no.~6, pp. 3109--3117, 2020.

\bibitem{6167509}
Y.~{Nam} and S.~{Hong}, ``Optimal placement of multiple visual sensors using
  simulation of pedestrian movement,'' in \emph{2012 International Conference
  on Computing, Networking and Communications (ICNC)}, 2012, pp. 67--71.

\bibitem{8335350}
A.~A. {Altahir}, V.~S. {Asirvadam}, N.~H.~B. {Hamid}, P.~{Sebastian}, N.~B.
  {Saad}, R.~B. {Ibrahim}, and S.~C. {Dass}, ``Optimizing visual sensor
  coverage overlaps for multiview surveillance systems,'' \emph{IEEE Sensors
  Journal}, vol.~18, no.~11, pp. 4544--4552, June 2018.

\bibitem{wirelessnetworkvisibility}
H.~Huang, C.-C. Ni, X.~Ban, J.~Gao, and S.~Lin, ``Connected wireless camera
  network deployment with visibility coverage,'' in \emph{Proceedings - IEEE
  INFOCOM}, 04 2013, pp. 321--322.

\bibitem{5170013}
Y.~{Yao}, C.~{Chen}, B.~{Abidi}, D.~{Page}, A.~{Koschan}, and M.~{Abidi}, ``Can
  you see me now? sensor positioning for automated and persistent
  surveillance,'' \emph{IEEE Transactions on Systems, Man, and Cybernetics,
  Part B (Cybernetics)}, vol.~40, no.~1, pp. 101--115, 2010.

\bibitem{Zhao2008}
J.~Zhao, S.-C. Cheung, and T.~Nguyen, ``Optimal camera network configurations
  for visual tagging,'' \emph{IEEE Journal of Selected Topics in Signal
  Processing}, vol.~2, no.~4, pp. 464--479, 2008.

\bibitem{zhao2009optimal}
J.~Zhao and S.~C. Sen-ching, ``Optimal visual sensor planning,'' in \emph{2009
  IEEE International Symposium on Circuits and Systems}.\hskip 1em plus 0.5em
  minus 0.4em\relax IEEE, 2009, pp. 165--168.

\bibitem{piecewise}
K.~L. Croxton, B.~Gendron, and T.~L. Magnanti, ``A comparison of mixed-integer
  programming models for nonconvex piecewise linear cost minimization
  problems,'' \emph{Management Science}, vol.~49, no.~9, pp. 1268--1273, 2003.

\bibitem{Nemhauser1988}
G.~L. Nemhauser and L.~A. Wolsey, ``Integer programming and combinatorial
  optimization,'' \emph{Wiley, Chichester. GL Nemhauser, MWP Savelsbergh, GS
  Sigismondi (1992). Constraint Classification for Mixed Integer Programming
  Formulations. COAL Bulletin}, vol.~20, pp. 8--12, 1988.

\bibitem{PyMIP}
\BIBentryALTinterwordspacing
Python-MIP, ``Mip optimizer for python,'' 2019. [Online]. Available:
  \url{https://python-mip.com}
\BIBentrySTDinterwordspacing

\end{thebibliography}

\end{document}